\pdfoutput=1

\documentclass[11pt]{article}
\usepackage{amsmath}
\usepackage[preprint]{acl}

\usepackage{times}
\usepackage{caption}
\usepackage{latexsym}
\usepackage{enumitem}
\usepackage{booktabs}
\usepackage[T1]{fontenc}
\usepackage{listings}

\usepackage{enumitem}
\usepackage{xcolor}

\definecolor{darkblue}{HTML}{000099}
\definecolor{codebg}{rgb}{0.95, 0.95, 0.95}

\usepackage[utf8]{inputenc}

\usepackage{microtype}

\usepackage{inconsolata}
\usepackage{booktabs}
\usepackage{array}
\usepackage{tabularx}
\newcolumntype{L}[1]{>{\raggedright\arraybackslash}p{#1}} 

\usepackage{makecell} 
\usepackage{graphicx}
\usepackage{appendix}

\usepackage{ragged2e}
\newcolumntype{Y}{>{\RaggedRight\arraybackslash}X}

\usepackage[most]{tcolorbox}
\tcbset{boxrule=0.4pt,arc=2pt, left=1mm,right=1mm,top=0.5mm,bottom=0.5mm}
\newtcolorbox{casebox}[1]{breakable, colback=white, colframe=black!20, title={#1}}

\title{Synthesizing Behaviorally-Grounded Reasoning Chains: A Data-Generation Framework for Personal Finance LLMs}

\author{Akhil Theerthala \\
  Perfios Software Solutions}

\begin{document}
\maketitle
\begin{abstract}
Personalized financial advice requires consideration of user goals, constraints, risk tolerance, and jurisdiction. Prior LLM work has focused on support systems for investors and financial planners. Simultaneously, numerous recent studies examine broader personal finance tasks, including budgeting, debt management, retirement, and estate planning, through agentic pipelines that incur high maintenance costs, yielding less than 25\% of their expected financial returns. In this study, we introduce a novel and reproducible framework that integrates relevant financial context with behavioral finance studies to construct supervision data for end-to-end advisors. Using this framework, we create a 19k sample reasoning dataset and conduct a comprehensive fine-tuning of the Qwen-3-8B model on the dataset. Through a held-out test split and a blind LLM-jury study, we demonstrate that through careful data curation and behavioral integration, our 8B model achieves performance comparable to significantly larger baselines (14-32B parameters) across factual accuracy, fluency, and personalization metrics while incurring 80\% lower costs than the larger counterparts. 

\end{abstract}

\section{Introduction}

Legal counseling, healthcare, and finance are among the numerous high-stakes domains in which personalized advice is essential. However, the development of this personalized advice is fraught with obstacles, requiring substantial investments and years of human expertise. Recent research efforts have thoroughly investigated automated decision support systems in various areas, emphasizing their cost-effectiveness. In the financial sector, a variety of support systems have been investigated, with a particular emphasis on asset recommendations and investment predictions. \cite{10.1007/978-3-031-56069-9_30, luo2025llm, 10.1145/3539618.3591850}

Recent advances in large language models (LLMs) have shown effective performance in acting as decision support systems for investors \cite{gupta2023gpt} and financial planners \cite{Huang_Che_Zheng_Li_2024}. The core advantage of natural language generation presents these automated support systems with a unique advantage that was never available in previous applications. This advantage has repeatedly shown its power in linguistic tasks such as streamlining complex financial narratives from extensive documents, corporate discourses, news sources, and social media.\cite{gueta2025llmslearnmacroeconomicnarratives, lee2024financewizardfinllmchallenge} The utility of these models is also being explored in Time series \cite{liu2025llm4ftsenhancinglargelanguage} and Financial reasoning applications \cite{liu2025finr1largelanguagemodel}.

Notwithstanding this capability, recent research indicates that no model excels across all financial task categories, which include text summarization, sentiment analysis, causal analysis, forecasting, and text classification \cite{matlin2025financelanguagemodelevaluation}. It has been demonstrated that attaining robust performance frequently necessitates the utilization of large, expensive models, thereby constraining the practicality of these solutions. Due to these inherent limitations and the complexity of financial advisory, many studies focusing on broader financial decision systems have preferred an agentic approach over training financial domain-specific language models. \cite{okpala2025agenticaisystemsapplied, joshi2025comprehensive, takayanagi2025generativeaiagentseffective}

Although the initial agentic frameworks focused on answering simple inquiries,\cite{10.1145/3604237.3626867} recent studies have accelerated the development of these systems to provide practical and actionable advice to the end user \cite{FinPersona,okpala2025agenticaisystemsapplied}. These agents can now dynamically interact with users and can assist in various tasks such as recommendation, question answering, search, and customer profiling. \cite{li2024agentframeworkrealtimefinancial,takayanagi2025generativeaiagentseffective,  han2024enhancinginvestmentanalysisoptimizing} 

Although agentic systems demonstrate potential in providing tailored financial advice, their efficacy is hindered by considerable constraints, including the integration with legacy systems, compliance with data security regulations, and high inference costs.  \cite{cemri2025multiagentllmsystemsfail, wang2025multiagentinteractivequestiongeneration}.  In support of these concerns, a recent study by \cite{meimandi2025measurementimbalanceagenticai} illustrates that a confluence of technical and cost-related factors hinders these applications from realizing even 25\% of their anticipated returns.  This research also establishes an important differentiation: success in benchmarks does not necessarily equate to success in deployment.  In practical terms, these proactive financial advisors frequently encounter a swift deterioration in performance within a matter of months following their implementation, attributable to the inherent volatility of real-world conditions.  Concurrently, studies show that the extent of personalization is often limited by the volume of context and information that can be supplied to an agent, impacting the overall performance. \cite{zhou-etal-2025-llms, biased_echoes}

One of the direct ways to address these limitations is to tune a model with a domain-specific context that integrates financial, behavioral, and psychological information. This work aims to close this gap by providing a reproducible framework to generate financial advice through a well-structured chain-of-thought. In particular, the framework constructs supervision data to train models to (a) provide personalized guidance for users' financial dilemmas, (b) reliably apply core financial knowledge, and (c) recognize and mitigate user-side behavioral biases by integrating behavioral and historical evidence.
 
To address these limitations, we propose a novel, data-centric framework for synthesising behaviorally-grounded reasoning chains. Rather than relying on complex agentic architectures, our approach directly bakes financial, behavioural, and psychological knowledge into the training data itself. Crucially, we treat the inference of the user's psychological state not as an afterthought, but as a standalone, foundational phase in the reasoning chain. This design choice is directly motivated by recent findings that users' trust and engagement are heavily influenced by the persona of the advisor \cite{takayanagi2025generativeaiagentseffective}, not just the raw accuracy of its advice. By isolating and explicitly modelling this psychological dimension, our framework ensures that personalisation and empathetic framing are intrinsic to the model's reasoning process, leading to more effective and trustworthy financial guidance. 

 It should also be considered that although recent agentic frameworks respond based on real-time knowledge; most of these knowledge sources need to be manually curated \cite{aggarwal2024overcoming}. In addition to this, we should note that most of the recommendations needed for general financial advice do not require real-time financial knowledge. Instead, this advice needs an agent that can inherently retrieve the relevant information from its memory. We address this problem by carefully crafting a chain-of-thought section to retrieve the financial context relevant to the query.

Recent studies have shown that inherent biases often limit users' ability to make many wealth-making financial decisions. \cite{baker2017behavioural, agrawal2012conceptual} These biases are highly variable and often depend on the age, experience and location of the user. Many financial agents do not directly address these biases when providing financial advice to the user. In this study, we have tried to integrate these biases into the reasoning model's natural chain-of-thought to tune the final responses towards acknowledging and addressing these biases.  

Each stage of chain-of-thought generation is verified by a set of Large Language Model juries that rank various generations and pick the best version suitable for the user queries. We used this framework to generate a 19k sample dataset, which is used to finetune a Qwen-3-8B model. This model is then compared to models of similar sizes to determine the impact of this framework.

This paper introduces a principled, data-centric framework as a step toward smaller, more trustworthy personal finance LLMs, and we outline its use as a backbone policy within agentic workflows to thin planning chains and lower orchestration cost—an evaluation we defer to future work.

\section{Related Works}

The application of automated systems to financial advice is not a new undertaking. Prior to the widespread adoption of large language models, research focused on applying classic techniques such as collaborative filtering and case-based reasoning to well-defined domains such as loan and insurance policy recommendation, as surveyed by \citet{zib_2016_rec_meets_fin}. However, the advent of powerful LLMs has opened new frontiers and presented a distinct set of challenges and approaches.

Much of the recent literature has focused on benchmarking the capabilities of general-purpose LLMs on a range of isolated financial tasks. For instance, a comprehensive study by \citet{Hean_2025} evaluated leading models such as ChatGPT and Claude against standardized financial literacy questionnaires covering diverse topics from mortgages to taxes. While their findings show that newer models are consistently improving and can achieve high accuracy on specific topics, they also reveal significant limitations, concluding that LLMs still struggle to provide accurate responses for complex financial queries. This highlights a critical performance gap: off-the-shelf models are often insufficient for the nuanced demands of holistic financial advice.

To overcome the limitations of single models and address more complex, multi-step planning, a significant body of research has shifted towards developing sophisticated agentic workflows. A recent survey by \citet{ding2024largelanguagemodelagent} provides a comprehensive overview of this landscape, categorizing these systems into distinct architectural patterns such as reflection-driven and debate-driven agents. A clear example is the work of \citet{okpala2025agenticaisystemsapplied}, who designed "agentic crews" composed of multiple specialized LLM agents, such as data scientists and compliance checkers, to automate the entire financial modelling and risk management pipeline. While powerful, such multi-agent systems demonstrate significant architectural complexity and high maintenance costs. Furthermore, research into these conversational agents has revealed significant risks; \citet{takayanagi2025generativeaiagentseffective} found in a user study that participants often placed more trust in a confident, "extroverted" agent even when it provided lower-quality advice, highlighting the potential for these complex systems to mislead inexpert users.

We argue, however, that the primary bottleneck is not architectural complexity, but the inherent irrationality of the models themselves, necessitating a data-centric approach. This need is rooted in the tendency of LLMs to amplify human cognitive biases. The groundbreaking work of \citet{zhou-etal-2025-llms} introduced a comprehensive framework based on behavioral finance to demonstrate that LLMs exhibit significant financial biases, such as anchoring and overconfidence. Their crucial finding that fine-tuning on financial data can sometimes exacerbate these irrational tendencies underscores the profound risks of using uncurated data. This is supported by empirical studies exposing a significant "product bias" in leading LLMs \cite{zhi2025exposingproductbiasllm} and by findings that LLM-generated advice systematically increases portfolio risk by reinforcing investment biases such as geographical concentration and trend chasing \cite{biased_echoes}. Taken together, these findings reveal that a model's pre-trained knowledge is an unreliable and potentially risky foundation for financial advice.

Therefore, our work addresses a critical gap. While large-scale financial language models like FinGPT, which continuously ingest real-time market data to update and adapt the underlying model \cite{yang2023fingpt, wang2023fingptbenchmark, zhang2023instructfingpt, 2023finnlp}, have been proposed, our approach differs fundamentally in its core contribution. Whereas such work focuses on scaling model capacity and live data ingestion, our work introduces a novel and reproducible methodology for creating the supervision data itself. By integrating the relevant financial context with behavioral finance studies, we construct a high-quality reasoning dataset designed to train smaller, more efficient end-to-end advisors that are grounded in sound, unbiased principles from their inception.

\section{Dataset construction}
\subsection{Data Collection and Processing}
\begin{table*}[htbp]
\small
\centering

\captionsetup{justification=raggedright,singlelinecheck=true}
\caption{\textbf{A detailed breakdown of the dataset generated via our proposed framework}. This table presents the distribution of approximately 19k samples across eight distinct categories of personal finance.  Each category includes key metrics, such as the average token count for the initial query, the generated chain-of-thought delineating the reasoning steps, and the final answer. }
\label{tab:pfcategories_raw}

\begin{tabular}{@{} l p{5cm} r r r r @{}}

\hline
\textbf{Category} & \textbf{Description} & \textbf{Count} &
\multicolumn{1}{c}{\textbf{\begin{tabular}[c]{@{}c@{}}Avg.\\Query\\Tokens\end{tabular}}} &
\multicolumn{1}{c}{\textbf{\begin{tabular}[c]{@{}c@{}}Avg.\\CoT \\Tokens\end{tabular}}} &
\multicolumn{1}{c}{\textbf{\begin{tabular}[c]{@{}c@{}}Avg.\\Response\\Tokens\end{tabular}}} \\
\hline

Debt Management \& Credit &
Strategies for debt reduction (e.g.\ snowball, avalanche), credit-score improvement, and loan analysis. &
5175 & 215.76 & 628.30 & 393.69 \\

Retirement Planning &
Strategies, income-needs analysis, benefits optimization (e.g.\ 401(k), pensions) and withdrawal strategies. &
3286 & 198.10 & 648.28 & 407.02 \\

Tax Planning \& Optimization &
Tax-minimization strategies, understanding deductions and credits, and investment-tax implications. &
3019 & 182.96 & 630.20 & 397.81 \\

Investing \& Wealth Building &
Investment strategies based on risk tolerance, diversification, asset allocation, and long-term growth. &
2994 & 200.16 & 653.54 & 402.98 \\

Budgeting \& Cash-Flow Management &
Creating budgets, tracking expenses, managing income streams, and improving cash flow. &
2503 & 221.53 & 628.71 & 394.47 \\

Insurance \& Risk Management &
Assessing insurance needs (life, health, property), understanding policies, and managing financial risks. &
1035 & 213.86 & 621.53 & 389.65 \\

Savings \& Emergency Funds &
Strategies for building savings, establishing emergency funds, and goal-based saving. &
638 & 177.18 & 652.25 & 382.95 \\

Estate Planning \& Legacy &
Wills, trusts, inheritance considerations, and minimising estate taxes (accounting for regional variations). &
196 & 216.90 & 653.47 & 409.06 \\
\hline
\end{tabular}
\end{table*}

Our first step was to collect a large pool of real-world finance questions. Reddit \cite{reddit} proved ideal as a source of complex scenarios that span the breadth of personal finance domains—from debt consolidation and retirement planning to tax optimization and insurance decisions. The platform's subreddits, particularly r/personalfinance, which receives hundreds of thousands to millions of user queries, contain authentic scenarios that capture the intricate, multi-faceted nature of real financial decision-making, providing the scenario diversity essential for training comprehensive advisory models.

To comply with Reddit's terms and conditions, we exclusively utilized publicly available archived data from posts prior to June 2023, ensuring all collected queries were ethically sourced and properly de-identified.

After ingestion, we filtered the raw corpus in two stages:
\begin{itemize}
    \item \textbf{Topical validity} – retained posts that contained an explicit, answerable personal finance question (e.g., budgeting, credit, retirement), discarding generic news, advertisements, or off-topic commentary. 
    \item \textbf{Contextual clustering} – grouped semantically similar posts and removed near-duplicates to reduce noise. 
\end{itemize}
    
This pipeline yielded 405k unique questions. We sampled 19k representative queries that span eight thematic categories. Table \ref{tab:pfcategories_raw} contains the detailed description of the final dataset generated using the framework. The entire 405k-item corpus remains available for future scaling. Details about prompt templates and specific instructions used in each phase of the generation framework are presented in \textbf{Appendix \ref{app:training_prompts}}.

\subsection{Generation methodology}
\label{sec:generation}

On a high level, the dataset generation framework can be divided into two parts: (i) chain-of-thought generation and (ii) response generation. 

\begin{figure*}

  \includegraphics[width=\textwidth]{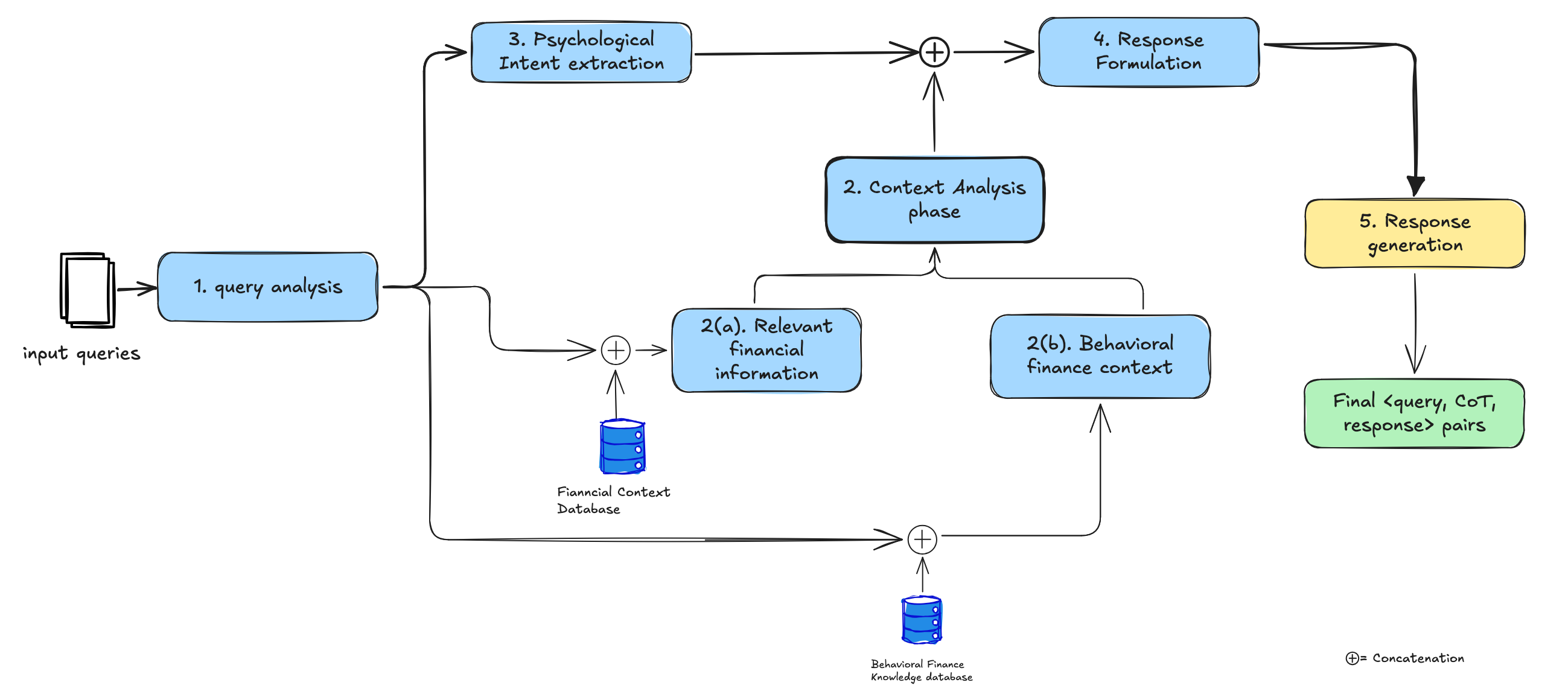}
  \captionsetup{justification=raggedright,singlelinecheck=true}
    \caption{Dataset generation pipeline. Four modular chain-of-thought phases feed into final response generation. Each phase includes LLM-jury validation (not shown) to ensure quality.}
  \label{fig:dataset_pipeline}
\end{figure*}

Our chain-of-thought generation is divided into four major phases, as illustrated in \textbf{Fig.~\ref{fig:dataset_pipeline}}. This modular approach helps us focus on developing an independent rubric for each phase while giving the ability to stitch them together as a coherent chain-of-thought. 

\subsubsection{Query Analysis}
The issue with natural language inquiries is the potential inconsistency of the information supplied to the model. There may be significant redundancy, or essential information may be hidden at times. Thus, the initial stage of answer creation, the question analysis phase, serves as a fundamental step in which the user's question is deconstructed into its essential components. This is required to ascertain the (i) primary conflict from the user's input; (ii) the principal players in the dilemma; and (iii) the essential financial facts to address the inquiry. This facilitates the optimization of subsequent cognitive processes while remaining aligned with the user's inquiry.

\subsubsection{Context Analysis}
\paragraph{Context analysis (Modular RAG).}
After intent parsing, we assemble a compact evidence pack via a modular RAG framework \cite{modular_rag} built on two self-curated corpora snapshotted through \textbf{February 2025}: (i) a \textbf{financial corpus} of $\sim$600k tokens—practical sources such as Investopedia and a Bogleheads snapshot \cite{investopedia_website,bogleheads_forum} covering core concepts (e.g., retirement accounts, debt-repayment strategies), plus curated summaries of policy changes for major U.S.\ credit-card products and other consumer-policy/market updates; and  (ii) a \textbf{behavioral corpus} of $\sim$300k tokens—research and practitioner write-ups spanning psychology of risk, investor behavior, behavioral portfolio theory, behavioral asset pricing, psychological effects of debt, and generational differences. 

Candidate chunks are retrieved with \texttt{text-embeddings-3-large} \cite{openaiembeddings} (top-25), re-ranked with \texttt{all-MiniLM-L12-v2} \cite{sentence_transformers}, and the top-15 are condensed by gemini-2.0-flash \cite{gemini2flash,geminiteam2025geminifamilyhighlycapable} to remove residual noise and unify terminology. 
The streamlined context and the user query then feed the downstream reasoning stage. Further details are provided in Appendix~\ref{app:modular-rag}.

\subsubsection{Psychological Cue identification}
In parallel to context identification, a psychological cue identification module is run to identify cues from the text. We extract the overall sentiment of the text, the primary emotions identifiable from the choice of words in the query, and the level of certainty present in the information. Using these cues, we try to generalize a set of communicative intents that might be behind the user's query. By breaking down the assessment into four distinct categories, the process ensures a comprehensive evaluation of the user's intent. This intent is utilized to direct the final response into a tone that is most suitable for the user, rather than directly providing them a monotonous response. 

To operate the cue-identification at scale, and in line with the prior studies which demonstrate that state-of-the-art large language models outperform human annotators in judgment tasks\cite{Bojić2025, DanielEAnalysis}, we adopt an LLM-based framework for cue identification similar to the other stages in the framework. 

\subsubsection{Response Formulation}
The final phase of the chain-of-thought is a distinct response formulation phase, in which we synthesize a set of instructions, consolidating information from all preceding phases. This produces a set of directives that must be adhered to throughout the response-generation phase.

\subsection{Response generation}
A conclusive response is formulated to address the user's inquiry, utilizing the previously optimized stages of information. This concluding comment is based on the financial context presented and is articulated in a suitable tone for the user.

\subsection{Data Validation}
Given that various open and proprietary LLMs automate numerous generations, there is a clear necessity to assess and authenticate their outputs. We employed a series of juries, specifically gemini-2.0-flash and o4-mini \cite{o4mini}, to evaluate and rank various generations for each phase. Each juror assessed the created information within a three-shot evaluation framework, ultimately selecting the highest-ranked response for subsequent generation jobs.

\section{Evaluation}
To test whether our dataset enables practical decision support, we fine-tune Qwen-3-8B \cite{yang2025qwen3technicalreport} for five epochs and compare it with baselines of similar size.

We perform an additional assessment of the performance using two separate held-out datasets. We employ these methods to assess the quality of the responses through both quantitative and qualitative measures. 

\subsection{Quantitative Evaluation}
To assess the quantitative performance of the models, we utilize a held-out dataset comprising 500 distinct queries across various categories of personal finance. Ground truths were produced by the generation framework presented in \textbf{Section \ref{sec:generation}} (not the fine-tuned model) prior to training and validated by independent jurors. Following the ground-truth generation, we calculate the BERTScore \cite{zhang2020bertscoreevaluatingtextgeneration} using the Qwen-3-8B-embeddings \cite{zhang2025qwen3embeddingadvancingtext} model to assess the semantic accuracy of the responses. We also calculate the BLEURT \cite{sellam2020bleurtlearningrobustmetrics} score to assess the fluency (or) human-likeness of the responses, respectively. The quantitative scores of various models utilized in this evaluation are detailed in \textbf{Table \ref{tab:auto_eval}}. 

\begin{table}[t]
\centering
\setlength{\tabcolsep}{6pt}

\begin{tabularx}{\linewidth}{>{\raggedright\arraybackslash}X cc}
\toprule
\thead{Model} & \thead{BERTScore $\uparrow$} & \thead{BLEURT} \\
\midrule
\makecell[l]{Gemma3-27B-IT \\ \cite{gemmateam2025gemma3technicalreport}}                & \textbf{0.7142} & 0.4374 \\
\makecell[l]{Gemma3-12B-IT}                & 0.7139          & 0.4390 \\
\makecell[l]{Mistral-24B-2501 \\ \cite{mistral}}  & 0.7133          & 0.4464 \\
\makecell[l]{QWQ-32B \cite{qwq32} \\ (reasoning)}    & 0.7069          & 0.4452 \\
\makecell[l]{DeepSeek-Qwen-14B \\ (reasoning)}\\ \cite{deepseek-r1-14b} & 0.7069          & 0.4513 \\
\makecell[l]{\textbf{Ours (8 B})} & 0.7000          & \textbf{0.4600} \\
Llama-3 8B \cite{llama8b}                                    & 0.6881          & 0.4547 \\
Mistral-7B v0.3 \cite{mistral7b}                            & 0.6650          & 0.4501 \\
\bottomrule
\end{tabularx}
\captionsetup{justification=raggedright,singlelinecheck=true}
\caption{Automatic evaluation on the 500-query test set. Bold marks the best score in each column; higher is better.}
\label{tab:auto_eval}
\end{table}

Our 8B model achieves semantic accuracy comparable to leading baselines, including Gemma3-27B/12B and Mistral-24B. In particular, our model surpasses these larger models by approximately 3–5\% in human-likeness and fluency. This indicates a reduced deviation from ground-truth data and enhanced fluency signals compared to models twice its size.

\subsection{Qualitative Evaluation}
\label{section-qualEval}

To complement reference-based metrics and, critically, to assess the model's generalization capabilities, we run a list-wise blind LLM-jury ranking on 504 queries that were entirely held out and unseen during the training phase. These test queries were collected from a subsequent time period to ensure no data contamination. Meanwhile, all the candidates were zero-shot generated in their respective default inference settings to get their best performance. This setup allows us to evaluate whether our fine-tuned model has merely learned to mimic the training data or if it has successfully internalized a generalizable framework for the response generation that can be applied to novel user problems. 

To mitigate familial bias and leakage, we excluded judges from model families used anywhere in our pipeline. In particular, Gemini models were omitted because they were used during dataset generation/validation, and Qwen-family judges were omitted because the system under test is Qwen-8B. A few otherwise suitable judges were also excluded for cost reasons. The final judge pool comprises models from unrelated families; none overlapped with training or data-creation components.

For each query, every judge sees all k anonymized candidates simultaneously (no ground truth and no model identities) and returns a full ranking; candidate order is uniformly randomized per replicate. We use two main judges, namely DeepSeek-V3-0324 \cite{deepseekai2025deepseekv3technicalreport} and Kimi-k2 \cite{moonshot_ai_2025}. Kimi-k2 is run three times, and DeepSeek-v3-0324 is run five times on independently shuffled anonymized candidate orders for each query to reduce possible biases. These judges were chosen in order to avoid same-family bias prevalent in modern LLM-judge studies.

\begin{figure}
    \centering
    \includegraphics[width=\linewidth]{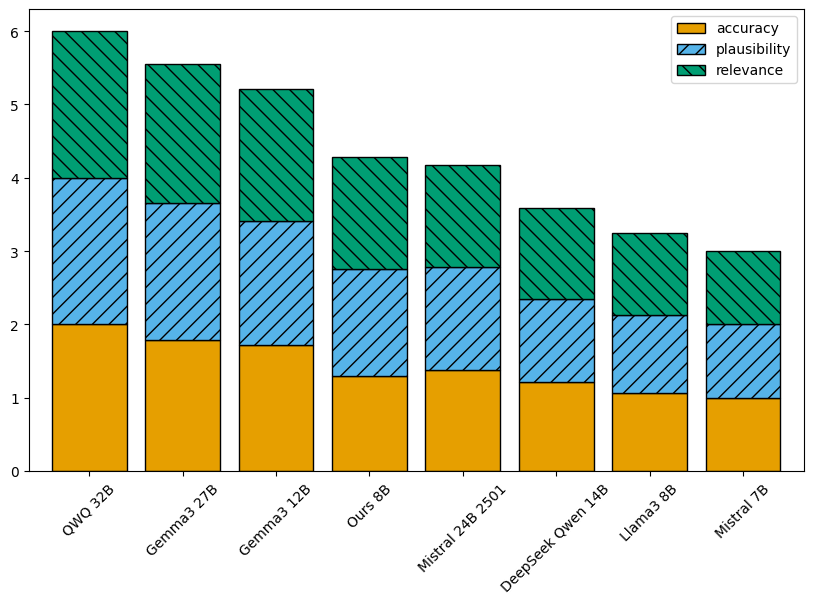}
    \captionsetup{justification=raggedright,singlelinecheck=true}
    \caption{LLM-jury evaluation on 504 unseen subreddit queries: stacked bars show Borda-average scores for accuracy (blue), plausibility (orange), and relevance (green); taller bars indicate stronger overall preference.  Our 8 B system (fourth from left) outperforms all other sub-14 B models and approaches the 27 B–32 B leaders. The y-axis represents the average Borda points a model has received.}
    \label{fig:overall_graph_trends}
\end{figure}

The rankings are converted to Borda points\cite{borda_count_2023} and averaged across judges and replicates to obtain the representative score of a response. We receive the ranking judgments according to three criteria, namely their financial accuracy, plausibility, and relevance to the query, and report the aggregate Borda scores in \textbf{Fig.\ref{fig:overall_graph_trends}}. Whereas \textbf{Appendix \ref{app:scores}} presents the in-depth analysis of the evaluation results. 

To examine rank consistency between the judge sets, we compute Kendall's $\tau$ and Spearman's $\rho$ over per-query model ranks.  Kendall's $\tau$ assesses pairwise order agreement (do both judges prioritize model A above model B?).  Spearman's $\rho$ assesses how closely the complete ranked lists move together and penalizes significant rank differences.  We observe $\tau\approx0.62$-$0.69$ and $\rho \approx 0.76$-$0.83$ (overall $\tau=0.64$, $\rho=0.79$), indicating substantial agreement. The consistently higher $\rho$ than $\tau$ suggests disagreements are mostly local swaps rather than wholesale reorderings.  Relevance demonstrates the strongest alignment ($\tau = 0.691$, $\rho = 0.826$). \textbf{Table \ref{tab:rank-corr}} shows $\tau$ and $\rho$ for each metric and overall.

\begin{table}[t]
\centering
\captionsetup{justification=raggedright,singlelinecheck=true}
\caption{Rank correlations between judge sets (higher is better). $\tau$ measures how often the judges agree with A > B, and $\rho$ measures how closely the full rank lists track.}
\label{tab:rank-corr}
\begin{tabular}{lcc}
\toprule
\textbf{Metric} & \textbf{Kendall's $\tau$} & \textbf{Spearman's $\rho$} \\
\midrule
Plausibility & 0.6183 & 0.7711 \\
Accuracy    & 0.6183 & 0.7635 \\
Relevance   & \textbf{0.6910} & \textbf{0.8264} \\
\midrule
Overall     & 0.6429 & 0.7904 \\
\bottomrule
\end{tabular}
\end{table}

Our experimental results demonstrate that a well-curated, behavior-tuned finance dataset can elevate an 8B open model to achieve performance parity with models two to three times its size, thus validating the practical utility of our framework. Details about the entire training environment and settings are presented in \textbf{Appendix \ref{app:training_details}}. 

\subsection{Qualitative Analysis and Error Patterns}

Analysis of the 504 held-out responses reveals consistent patterns across the three evaluation dimensions. When models produce inaccurate responses, they typically also exhibit degraded reasoning quality—accuracy and plausibility failures often co-occur. However, relevance remains relatively independent; responses can stay on-topic and address user constraints even when containing factual errors or poor reasoning.

\textbf{Strengths.} The model consistently produces well-structured responses with clear headers, sequential action steps, and appropriate empathetic framing. It reliably extracts user-specific details (monetary amounts, timelines, constraints) and incorporates them into tailored advice. Responses typically acknowledge emotional context before providing practical guidance—a pattern that enhances perceived helpfulness.

\textbf{Failure Modes.} The primary weakness is factual hallucination, particularly for jurisdiction-specific regulations and tax details. The model occasionally generates plausible-sounding but incorrect specifics (e.g., non-existent grant programs, outdated tax brackets). These errors are most frequent in regulation-heavy domains (taxes, insurance) and least common in general planning tasks (budgeting, debt management).

\textbf{Implications.} While the model maintains strong structural and empathetic qualities across all responses, factual grounding remains the key bottleneck. This suggests that adding targeted retrieval for regulatory information and calculation verification would yield the highest marginal improvement. Even with current limitations, the model's consistent task alignment and user-responsive framing provide practical utility for non-critical advisory scenarios.

\subsection{Cost Analysis}

\begin{table*}[htbp]
\centering

\captionsetup{justification=raggedright,singlelinecheck=false} 
\caption{
    Cost and Inference Performance Analysis for Deployment. Total costs reflect the expense to 
    infer 504 queries from the test set, with each model benchmarked using four concurrent requests.
}
\label{tab:cost_analysis}

\begin{tabular}{{@{} l c c c c c c @{}}} 
\toprule
\textbf{Model} & \textbf{Size (GB)} & \textbf{Endpoint Cost} & \textbf{GPU} & \textbf{Inference Time} & \textbf{Total Time} & \textbf{Total Cost} \\
 & & \textbf{(\$/h)} & & \textbf{(s/query)} & \textbf{(h)} & \textbf{(\$)} \\
\midrule
QWQ-32B & 65.0 & 3.8 & 4xL4 & 167.86 & 5.82 & 22.33 \\
Gemma3-27B & 46.4 & 2.5 & 1xA100 & 64.34 & 2.23 & 5.63 \\
Gemma3-12B & 20.0 & 1.8 & 1xL40S & 58.26 & 2.02 & 3.67 \\
\textbf{Ours (8B)} & \textbf{16.4} & \textbf{0.8} & \textbf{1xL4} & \textbf{34.15} & \textbf{1.19} & \textbf{0.96} \\
Mistral-24B-2501 & 46.1 & 3.8 & 1xA100 & 37.99 & 1.32 & 5.05 \\
DeepSeek-Qwen-14B & 29.5 & 1.8 & 1xL40S & 54.18 & 1.88 & 3.41 \\
Llama3-8B & 16.1 & 0.8 & 1xL4 & 33.58 & 1.17 & 0.94 \\
Mistral-7B & 14.5 & 0.8 & 1xL4 & 29.15 & 1.01 & 0.82 \\
\bottomrule
\end{tabular}
\end{table*}
Beyond performance metrics, practical deployment requires careful cost consideration. \textbf{Table \ref{tab:cost_analysis}} presents a comprehensive cost analysis of the model produced by our framework against several baselines, comparing hosting infrastructure, inference latency, and total operational expenses.

Our data-centric approach delivers exceptional cost efficiency in the personal finance domain. By enabling a compact 8B model to achieve performance competitive with much larger systems, our method facilitates at least an 80\% reduction in operational costs when compared to baselines with over 12B parameters. This dramatic cost reduction stems from targeted behavioral integration and principled data construction, rather than sheer computational scale.

The efficiency translates to practical deployment advantages: at a hosting cost of \$0.8 per hour and an average inference time of 34.15 seconds, our model enables responsive financial advisory services without prohibitive infrastructure requirements.  These results validate the effectiveness of our novel data generation framework. They demonstrate that by carefully integrating financial and behavioral signals into training data, it is possible to create competent, domain-specific models that are also economically viable. This presents a compelling approach for developing production-ready financial advisory tools that do not rely solely on expensive, large-scale models.

\section{Future Works}
We will advance this research by first determining the optimal path for global scaling: either finalizing a US-optimized pipeline for systematic market porting or—contingent on high-precision detection of regional signals (e.g., currency symbols, policy terminology, and spelling conventions)—implementing a Mixture-of-Experts (MoE) framework. In the latter case, a shared backbone model will process universal financial logic while lightweight regional experts handle localized nuances. This core model will deploy as a backbone policy within a thin agentic stack, minimizing latency and cost by resolving queries internally and invoking external tools (e.g., regulatory databases or fact-checking APIs) only for uncertainty resolution. We will rigorously measure resulting cost-latency trade-offs across regions. Rather than additional supervised fine-tuning, we will treat financial advice generation as an alignment problem, testing preference-based optimization (e.g., DPO/IPO) to refine outputs and deploying rule-based compliance layers to enforce regulatory fidelity, bias mitigation, and tone consistency. Success will be quantified through targeted evaluations of safety, compliance adherence, and user trust metrics.

\section{Conclusion}
Our research establishes a data-centric framework that enables an 8B-parameter model to achieve semantic fidelity and human-likeness on par with, and sometimes exceeding, 27–32B baselines in our held-out evaluations and blind LLM-jury study. On a 500-query test, the model outperforms Gemma3-27B by ~5\% on BLEURT and is competitive on BERTScore, with only a ~2\% difference; jury rankings show the 8B system approaching the 27–32B leaders. These gains stem from three synergistic components: explicit psychological cues, retrieval-augmented grounding, and a thin agentic execution layer. The modular design supports incremental extension (e.g., regional experts with minimal retraining). While geographic scope, behavioral depth, and privacy safeguards remain limitations, this work offers a cost-aware backbone for stand-alone personal-finance assistants and a viable alternative to monolithic cloud deployments—leaving a precise cost/latency audit to future work.

\section*{Limitations}
Several aspects of our work leave room for future improvements. First, our study is limited to inquiries sourced solely from Reddit, which may overlook other demographics and query formats, suggesting a need for more diverse data sources. Second, our 19k sample dataset, though sufficient for proof-of-concept, lacks the scale and diversity needed to cover the full spectrum of real-world personal finance scenarios. Future work should expand the corpus with varied sources beyond Reddit to improve generalization. Third, our psychological analysis remains rudimentary, deriving only basic sentiment from phrases rather than incorporating enhanced psychological indicators such as risk tolerance or financial stress through specialized surveys or transfer learning from clinical datasets. Finally, our framework's scope excludes tasks beyond core natural language processing, particularly multi-modal data processing and reasoning capabilities, which represent critical areas for future research expansion.

\section*{Acknowledgements}
I want to express my sincere gratitude to \textbf{Raghu Ram Theerthala (KPIT Technologies)} for his valuable contributions to the related works section and insightful discussions during the brainstorming sessions that helped shape this research. I am grateful to Prathyusha Akundi, Syed Md. Bilal, Ashish Kubade, and Sai Narayan for their careful review of the manuscript and constructive feedback that improved the clarity and quality of this work. This research was supported by Perfios Software Solutions, which sponsored the computational costs and infrastructure required for model training and evaluation.

\section*{Data \& Code Availability}
The dataset, model, and code artifacts described in this paper are publicly available on Hugging Face. All data has been de-identified following the ethical guidelines described in Section 6, with personally identifiable information removed from Reddit sources. The resources are released under the Apache 2.0 license to facilitate reproducibility and future research in behavioral finance and LLM applications.

The following resources are available:
\begin{itemize}
    \item \textbf{Model}: Fine-tuned Qwen-3-8B model at \\
    \url{https://huggingface.co/Akhil-Theerthala/Kuvera-8B-qwen3-v0.2.1}
    
    \item \textbf{Dataset}: 19k sample reasoning dataset at \\
    \url{https://huggingface.co/datasets/Akhil-Theerthala/Kuvera-PersonalFinance-V2.1}
\end{itemize}

\section*{Ethical Considerations}
\label{sec:ethics}
We curate data from publicly available Reddit posts and aggressively de-identify them: usernames/links/metadata are removed, PII (e.g., names, emails, phone numbers, addresses, IDs) is scrubbed, and queries are lightly rephrased so only the financial situation remains; no raw identifiers are stored or released. The system is for educational use only—not fiduciary or personalized financial advice—and our prompts/filters forbid unsafe guidance (e.g., evasion, “guaranteed returns”). Evaluation uses multiple LLM judges; we report inter-judge agreement and run judge-swap checks to limit model-family bias.

\bibliography{main}

\appendix
\appendixpage
\section{Prompting Guidelines followed for the generation and evaluation stages}

\subsection{Guidelines followed in the generation stage.}
\label{app:training_prompts}
This section focuses on outlining the guidelines followed in crafting the prompts for each phase of generating and evaluating the outputs.

\subsubsection{Overarching principles}
There are three core principles followed for the process of crafting the prompts:
\begin{enumerate}[label=\alph*.]
    \item Modularity
    \item Deconstruction
    \item Personification
\end{enumerate}

The goal of the overall prompt crafting is to keep the overall structure of the prompts similar and swappable depending on the task at hand. As with the framework, where the complex task of generating a suitable response is broken down into individual phases, the prompts are broken down to make sure the structure of the instructions given to the model remains the same.

Each stage of the prompting had a unique, suitable persona (e.g., linguistic analysis expert, expert financial reasoning engine). This role-playing technique primes the model to access relevant knowledge, adopt the appropriate tone, and constrain its behavior to the specific requirements of the task.

The generic structure of the prompt is as follows:
\begin{lstlisting}[language=Python]
"""
You are a {persona}, whose task is to {task_details}.

### INSTRUCTION ###
{instructions_for_the_task}

### Key Points ###
{key_points_to_keep_in_mind}

---
**Inputs**:
{inputs}
---
**Your Response**:"""
\end{lstlisting}

\subsubsection{Individual Phases}
\begin{enumerate}
    \item \textbf{Classification:}
    \begin{enumerate}[label=\alph*.]
        \item The primary goal of this stage is to classify incoming user queries into suitable categories of personal finance. The prompt constrains the model by forcing a single-label classification (ONE of the following) based on PRIMARY INTENT, which prevents ambiguity and ensures a decisive output for downstream routing.
        \item Each category has a Scope and an example that the model uses to make its decisions. If the query does not fall into any of the categories, the query is labeled Not\_Applicable.
    \end{enumerate}

    \item \textbf{Query Analysis:}
    \begin{enumerate}[label=\alph*.]
        \item The primary goal of this prompt is to direct the model to break down the user query into more specific and manageable pieces of information.
        \item Since most of the user queries on Reddit and in general are often filled with unrelated noise, this stage directs the model to distil the user's query into essential semantic elements, eliminating the conversational distractions and concentrating on actionable concerns and their impact on the key stakeholders.
    \end{enumerate}

    \item \textbf{Context Analysis:}
    \begin{enumerate}[label=\alph*.]
        \item The context analysis is one of the key prompts that influences the quality of the output by the framework. The prompt directs the final model to generate actionable and insightful contextual summaries that are placed into the model's natural chain-of-thought.
        \item The prompt explicitly asks for a Concise chain-of-thought Analysis Block and instructs the model that this is an internal reasoning step, not the final answer. This step forces the model to externalise its reasoning process, exploring multiple scenarios and their consequences before concluding.
        \item By requiring the model to detail the Stakeholder Impact for each approach, the prompt ensures a holistic analysis that considers the financial and emotional consequences for all relevant parties mentioned in the query. This scenario-based analysis moves the responses beyond simple fact-based analysis to a more human-centred form of reasoning.
    \end{enumerate}

    \item \textbf{Psychological analysis}
    \begin{enumerate}[label=\alph*.]
        \item The goal of this prompt is to direct the model and extract the key information about the user's state of mind when asking the query.
        \item The prompt demands that every conclusion about sentiment, emotion, or intent be justified by referencing specific words or phrases. This approach grounds the analysis in textual evidence, preventing the model from making unfounded psychological assumptions and improving the explainability of its affective understanding.
        \item This analysis is a separate step from the financial reasoning (Context Analysis). This deliberate separation prevents the user's emotional state from biasing the objective financial analysis, and vice-versa, allowing for a final response that can synthesise both aspects without compromising either.
    \end{enumerate}

    \item \textbf{Response Rubric}
    \begin{enumerate}[label=\alph*.]
        \item This stage consolidates all the previously collected information and creates a complete rubric that can direct the model into generating the final response.
        \item The key information from the previous stages gets highlighted while being linked to different parts of the user query for easier reference and understanding.
    \end{enumerate}

    \item \textbf{Response Generation}
    \begin{enumerate}[label=\alph*.]
        \item This final stage synthesises all preceding analyses into a coherent, user-facing response.
        \item The prompt provides the model with all previous outputs (the original query and the comprehensive chain-of-thought) and explicitly instructs it to integrate both factual accuracy and emotional intelligence seamlessly. It acts as a final "assembly" instruction, guiding the model on how to combine the rational and affective components.
        \item The use of clear positive (Do) and negative (Do not) instructions creates strict behavioral boundaries. For instance, "Do not reference the chain-of-thought analysis" ensures the final output is natural and user-friendly, hiding the complex underlying cognitive architecture from the end-user. These instructions create a helpful response without being robotic or transparent about its inner workings.
        \item These responses are generated in a way that ensure the ability to train non-reasoning models from the same dataset.
    \end{enumerate}
\end{enumerate}

\subsection{Prompt Guidelines for Evaluation through LLM-as-a-Judge}

The goal of the evaluation is to determine which responses are naturally ranked better than the others. Since this is a list-wise ranking with a high room for confusion or hallucination, the evaluation criterion are strictly defined. 

The overall prompt structure for each of the case are as follows:

\begin{lstlisting}[language=Python]
    
"""
You are a {persona}. Your task is to rank financial advice responses from best to worst based *solely* on the strict definition of {target_aspect}.

### **Evaluation Criteria**
{Evaluation Criterion}

#### **I. Primary Criteria (What to look for):**
{primary_set_of_instructions}

#### **II. Explicit Penalties (What to penalize):**
{penalizing_instructions}

#### ** III. Key Points to note:**
{additional_instructions}
---

**Query:** {query}

**Responses to Rank:**
{anonymized_shuffled_model_responses}
"""
\end{lstlisting}

\begin{enumerate}
    \item \textbf{Accuracy}:
    \begin{enumerate}[label=\alph*.]
        \item The goal of this prompt is to direct the model to review the search results and the query to estimate the accuracy of the output.
        \item The responses are penalized if and only if the responses demonstrate wrong/harmful advice (or) inappropriate financial concepts to the query. 
        \item The model is specifically instructed not to penalise on the style or relevance of the response and solely focus on the accuracy of the financial concepts provided in the text. This guides the model to rank solely based on the  accuracy of the financial concepts present in the response. 
    \end{enumerate}
    \item \textbf{Plausibility}:
    \begin{enumerate}[label=\alph*.]
        \item A response is defined to be plausible if it sounds reasonable and believable to a typical user. Some of the key characteristics include
        \begin{itemize}
            \item Logical flow and coherent reasoning structure
            \item Sensible approach to the problem
        \end{itemize}
        \item A response is penalized if it contains unnecessarily verbose or contain excessive detail. The responses are also penalized if they contain complex or hard-to-follow reasoning.
        \item The model is specifically instructed not to penalise on the accuracy or relevance of the responses. 
    \end{enumerate}
    \item \textbf{Relevance}:
    \begin{enumerate}[label=\alph*.]
        \item A response is considered relevant if it address every component of the user's query. A relevant response should incorporate the specific figures, constraints, and details mentioned in the user's query, and answer the questions immediately without generic introductions. 
        \item Any partial relevance or additional context not relevant to the query is penalized.
    \end{enumerate}

\end{enumerate}

\section{Modular RAG for Context Analysis}
\label{app:modular-rag}

\paragraph{Goal.}
Given a user query, the context-analysis phase assembles a compact, high-signal context pack from two specialized corpora:
(i) \emph{Behavioral insights} (behavioral economics and psychology) and
(ii) \emph{Financial concepts} (mainstream personal finance knowledge).
The context pack is then passed to the response generator.

\paragraph{Corpora.}
\textbf{Behavioral insights} are sourced from peer-reviewed research and reputable psychology venues, complemented by carefully selected psychology blogs for practitioner framing.
\textbf{Financial concepts} are drawn from practical, high-visibility sources such as Investopedia, Bogleheads, and other widely cited personal-finance viewpoints.
All raw pages are converted to Markdown with headers and section structure preserved to retain document semantics.

\paragraph{Preprocessing and indexing.}
\begin{itemize}[leftmargin=*]
  \item \textbf{Scraping \& normalization:} We scrape public pages (respecting robots/terms), remove boilerplate (nav, ads), and normalize to Markdown with stable headings.
  \item \textbf{Semantic chunking:} Documents are segmented into \emph{modular chunks} along header/semantic boundaries to keep each chunk topically coherent; we attach metadata (source, URL or handle, snapshot time, section path, corpus tag: \texttt{behavioral} or \texttt{financial}).
  \item \textbf{Dense indexing:} Each chunk is embedded with \texttt{text-embeddings-large-003} and stored in a vector databsase (ChromaDB).
\end{itemize}

\paragraph{Retrieval and re-ranking (per query).}
\begin{enumerate}[leftmargin=*]
  \item \textbf{Dual retrieval:} From each index, retrieve the top-$k$ candidates ($k{=}25$) using the query embedding.
  \item \textbf{Cross-encoder re-ranking:} Concatenate candidates from both corpora and re-rank with a lightweight cross-encoder (\texttt{sentence-transformers/all-minilm-l12-v2}); keep top-$m$ ($m{=}15$).
  \item \textbf{LLM synthesis/filter:} A fast LLM (gemini-2.0-flash) receives \{top-$m$ chunks, query\} and (a) extracts salient facts, definitions, and decision criteria; (b) discards residual off-topic spans; (c) emits a streamlined, source-attributed context.
\end{enumerate}

\paragraph{Assembly and handoff.}
The streamlined context (with inline source attributions and corpus tags) is passed, together with the user input, to the final LLM that completes the context-analysis phase.

\paragraph{Behavioral vs. financial module roles.}
The \textbf{behavioral module} surfaces cognitive-bias descriptors, debiasing tactics, and user-state cues (e.g., loss aversion framing, present bias prompts).
The \textbf{financial module} surfaces actionable rules of thumb, definitions, procedures, and typical constraints (e.g., contribution limits, insurance concepts, payoff ordering heuristics).
Both modules contribute to the same context pack; behavioral cues guide \emph{how} advice is framed, while financial chunks ground \emph{what} advice is provided.

\paragraph{Limitations.}
(1) Coverage and staleness depend on the snapshot of public sources; (2) blogs can introduce style bias despite re-ranking; (3) the synthesis step may over-prioritize well-structured sources.
We mitigate these by preserving source attributions, tracking snapshot timestamps, and prompting synthesis to prefer higher-priority sources when conflicts arise.

\section{Deeper Evaluation Results}
\label{app:overall_eval}
\subsection{Score Definitions and Rationale}
\label{app:scores}
We evaluate responses along three orthogonal axes—\textit{Accuracy}, \textit{Plausibility}, and \textit{Relevance}—to separate factual correctness, reasoning quality, and task alignment. This decomposition avoids a single scalar that can reward fluent but unsafe answers or penalize terse yet correct ones, and it enables targeted error analysis and ablations. 

\paragraph{Accuracy (financial correctness).}
\begin{itemize}[leftmargin=*, itemsep=2pt, topsep=2pt]
  \item \textbf{Objective.} Judge reviews the response against the query and retrieved evidence and scores only the \emph{validity of financial concepts, calculations, and advice}.
  \item \textbf{Penalties.} Deductions occur \emph{iff} the answer contains wrong or harmful guidance, or misapplies financial concepts to the user’s situation.
  \item \textbf{Non-considerations.} Style, tone, verbosity, and even partial coverage are \emph{not} penalized; the judge is instructed to focus exclusively on correctness.
\end{itemize}

\paragraph{Plausibility (reasoning quality).}
\begin{itemize}[leftmargin=*, itemsep=2pt, topsep=2pt]
  \item \textbf{Objective.} Assess whether the answer reads as reasonable and believable to a typical user—i.e., it exhibits a clear logical flow and a coherent problem-solving structure.
  \item \textbf{Penalties.} Overly verbose, needlessly complex, or hard-to-follow chains of reasoning are penalized.
  \item \textbf{Non-considerations.} Factual correctness and topical coverage are \emph{not} scored here; the lens is purely rhetorical/structural.
\end{itemize}

\paragraph{Relevance (task alignment).}
\begin{itemize}[leftmargin=*, itemsep=2pt, topsep=2pt]
  \item \textbf{Objective.} Verify that the response directly addresses \emph{every} component of the user’s query, incorporates the user’s numbers, constraints, and context, and answers without generic preambles.
  \item \textbf{Penalties.} Partial coverage, tangential content, or extra context not pertinent to the query is penalized.
  \item \textbf{Non-considerations.} Factual accuracy and stylistic polish are ignored for this axis.
\end{itemize}

\subsection{Borda Points}

\textbf{Definition.}  
For a listwise ranking of \(n\) systems, the item placed at rank \(r\) (\(r=1\) is best) receives a Borda score
\[
b = n - r,
\]
so the top entry gets \(n-1\) points and the last gets \(0\).

\textbf{Motivation.} Borda aggregation is well–suited to LLM‐as‐a‐judge experiments where \emph{relative} quality matters more than absolute scores:  

\begin{itemize}
    \item \emph{Full-order utilisation}: every position contributes signal, ensuring that small but consistent advantages are captured rather than discarded by winner-takes-all rules.  
    \item \emph{Cardinal comparability}: with a fixed candidate set, raw points can be averaged across queries and judges without normalisation, giving a stable, interpretable mean.  
    \item \emph{Robustness to mild noise}: swapping adjacent middle ranks changes the total by only \(\pm1\), so individual judge idiosyncrasies exert limited influence on the final average.  
\end{itemize}

\textbf{Interpretation.}  
Higher mean Borda points indicate that a system outranks its peers more often. The maximum possible mean is \(n-1\); the gap to this ceiling offers an intuitive sense of head-room.

\textbf{Limitations.}  
\begin{itemize}
    \item \emph{Rank-reversal}: inserting or removing a candidate can change every system’s score, complicating longitudinal comparisons.  
    \item \emph{Independence of Irrelevant Alternatives (IIA) violation}: a judge’s relative preference between two systems can affect, and be affected by, ranks assigned to others.  
    \item \emph{Equal-interval assumption}: the method treats the gap between successive ranks as uniform, ignoring situations where judges perceive larger quality jumps near the top.  
    \item \emph{Strategic susceptibility}: if human judges know what influences the aggregation, they could inflate or deflate lower ranks to benefit a favored system.  
\end{itemize}

\subsection{LLM-Jury Protocol}
\label{app:llmjury}
LLM-based judging scales across topics, is inexpensive, and achieves strong agreement with human raters when rubrics are explicit and task context is provided. It also captures holistic qualities (e.g., coherence, task fit) that single-number similarity metrics may miss.

It should be noted that  zero-shot judging is vulnerable to \emph{position bias} (earlier items rank higher), \emph{same-family bias} (preference for outputs from the judge’s own family), and prompt/leniency variance. We therefore (i) use \emph{multi-shot} prompts to anchor criteria, (ii) evaluate with \emph{listwise} ranking on \emph{independently shuffled} candidate lists, and (iii) diversify judges across model families to minimize correlated bias.

\paragraph{Judge pool and prompting.}
We employ two main heterogeneous judges: \textit{DeepSeek-v3-0324} (5-shot), \textit{Kimi-k2} (3-shot). For each query and criterion (Accuracy, Plausibility, Relevance), judges rank anonymized model outputs in a single list. Few-shot exemplars are held constant within a run and varied across repeats to reduce overfitting to any one demonstration set. A subsample of these rankings were further validated by \textit{o4-mini} model to consolidate the relative performance. 

\paragraph{Scoring and aggregation (per criterion).}
For each query, judges perform \emph{multi-shot listwise ranking} over anonymized outputs using the rubrics in Sec.~\ref{app:overall_eval}. Ranks are converted to raw Borda points $b = n - r$. We then:
\begin{enumerate}[leftmargin=*, itemsep=2pt]
  \item average $b$ across shuffles/repeats for each judge;
  \item average across the judges to obtain a per-query, per-criterion score for each model;
  \item average across all queries within a \emph{category} (e.g., the “overall” set or a PF subcategory) to obtain the model’s \emph{criterion-wise mean} in that category.
\end{enumerate}
The stacked bars in Fig.~\ref{fig:overall_graph_trends} display these criterion-wise means (\emph{Accuracy}, \emph{Plausibility}, \emph{Relevance}) for each model. For a single category-level number, we also report the \emph{unweighted average} of the three criterion-wise means as the model’s final representation score in that category.

\subsection{Overall Category Scores (Accuracy, Plausibility, Relevance)}
\label{app:overall-cats}

We report criterion-wise means derived from the raw Borda points assigned by the LLM jury (Sec.~\ref{app:llmjury}). For each criterion and model, scores are averaged across judges and queries within the overall set. Higher is better.

\begin{figure}[t]
  \centering
  \includegraphics[width=\columnwidth]{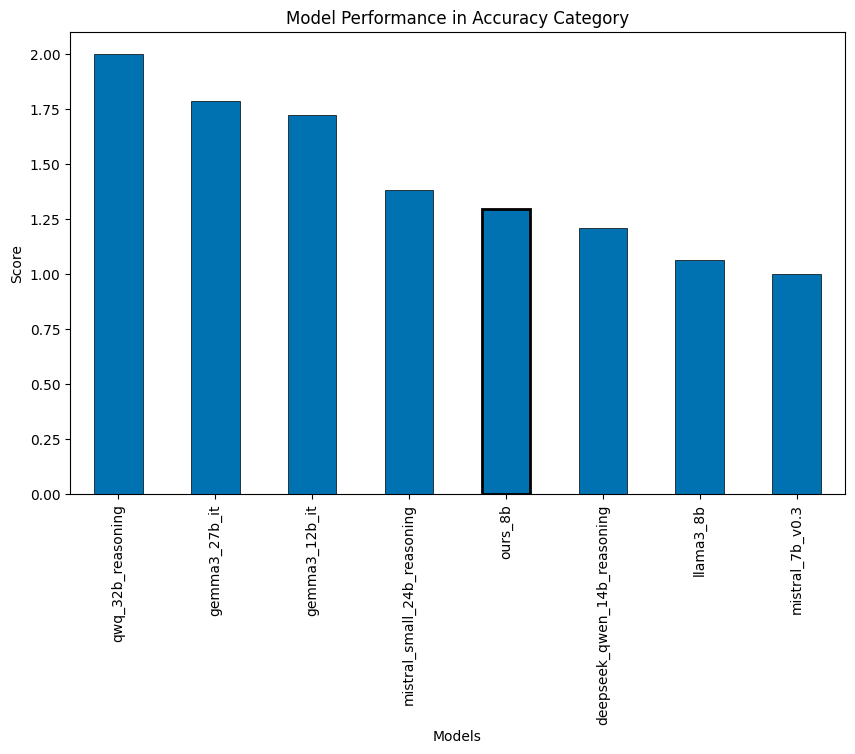}
  \captionsetup{justification=raggedright,singlelinecheck=true}
  \caption{Accuracy (mean raw Borda points per query, averaged over judges). A size-driven lead is visible; the proposed 8B is mid-pack, indicating factual calibration as the primary improvement lever.}
  \label{fig:acc-overall}
\end{figure}

\paragraph{Accuracy.}
Figure~\ref{fig:acc-overall} shows a size-tilted pattern: \textit{QwQ-32B (reasoning)} leads, followed by \textit{Gemma3-27B-it} and \textit{Gemma3-12B-it}. \textit{Mistral-Small-24B} sits between this top cluster and the rest. The proposed \textit{8B} model is mid-pack—behind the leaders and the 24B baseline, but ahead of several 7--14B baselines. This points to factual calibration and retrieval/verification as the primary levers to close the gap, rather than rewriting or stylistic tuning.

\begin{figure}[t]
  \centering
  \includegraphics[width=\columnwidth]{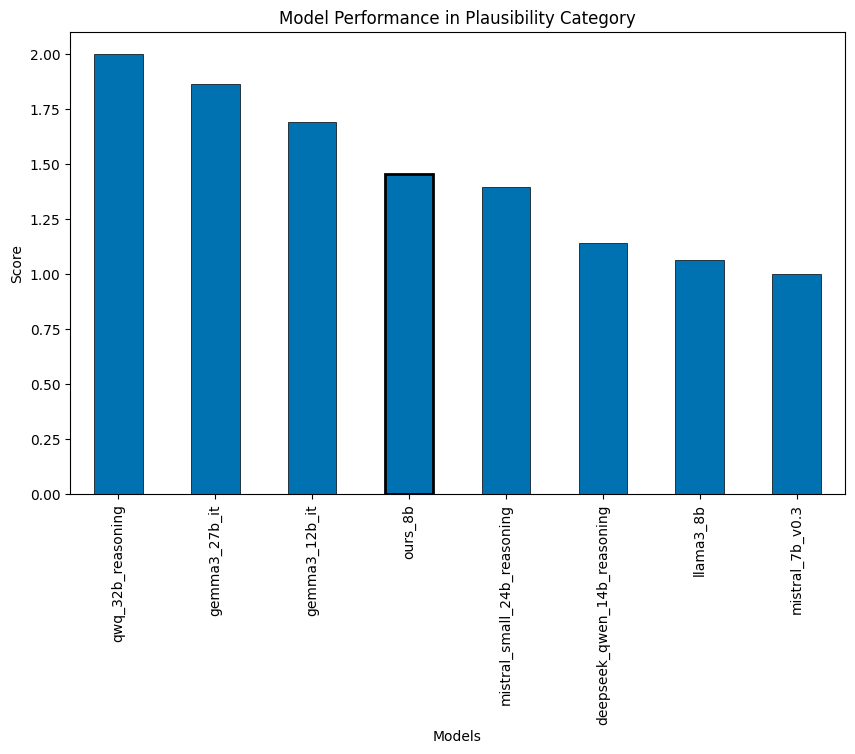}
  \captionsetup{justification=raggedright,singlelinecheck=true}
  \caption{Plausibility (mean raw Borda points). The proposed 8B clusters near the front and matches or exceeds several larger baselines, reflecting strong logical flow and coherent reasoning.}
  \label{fig:plaus-overall}
\end{figure}

\paragraph{Plausibility.}
As shown in Fig.~\ref{fig:plaus-overall}, \textit{QwQ-32B} ranks first, with \textit{Gemma3-27B-it} next. The proposed \textit{8B} clusters near the front: it exceeds the \textit{Mistral-Small-24B} baseline but trails \textit{Gemma3-12B-it}. This suggests that the dataset structure and few-shot conditioning induce coherent reasoning steps and a sensible flow even at mid scale.

\begin{figure}[t]
  \centering
  \includegraphics[width=\columnwidth]{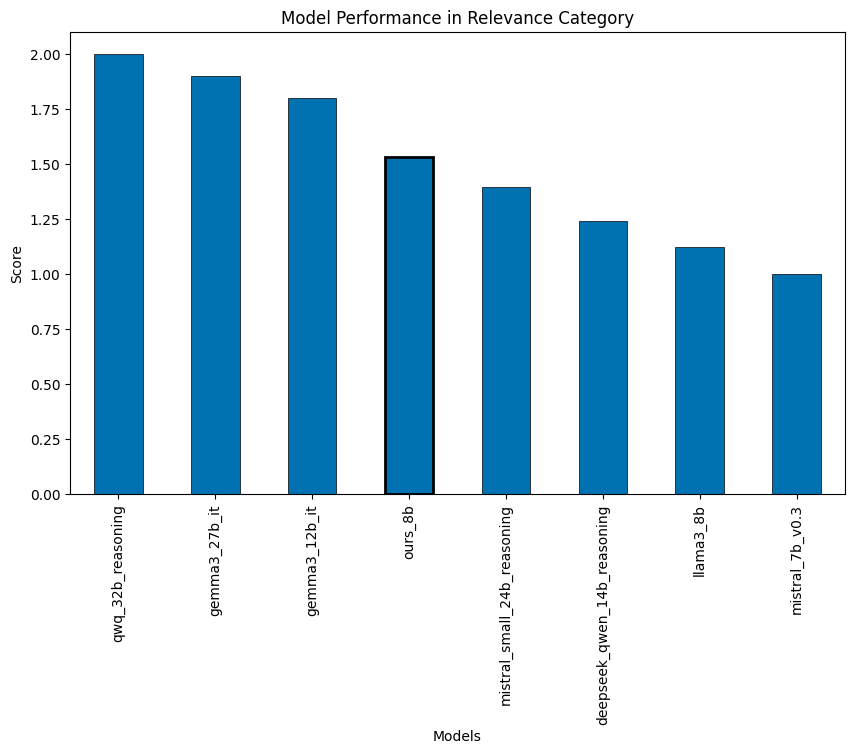}
  \captionsetup{justification=raggedright,singlelinecheck=true}
  \caption{Relevance (mean raw Borda points). The proposed 8B ranks immediately behind the top three, ahead of other baselines, indicating consistent mapping from user constraints to concrete answers.}
  \label{fig:rel-overall}
\end{figure}

\paragraph{Relevance.}
Figure~\ref{fig:rel-overall} indicates strong task alignment at the top end (\textit{QwQ-32B}, \textit{Gemma3-27B-it}, \textit{Gemma3-12B-it}). The proposed \textit{8B} ranks next (4/8), ahead of the remaining baselines, suggesting it reliably maps user constraints and addresses all parts of the query without drifting into generic preambles. The residual gap likely reflects cases that require exhaustive edge handling (e.g., niche eligibility rules) rather than broad intent recognition.

\paragraph{Cross-criterion takeaway.}
Across criteria, the proposed \textit{8B} model is \emph{plausibility}– and \emph{relevance}-competitive while lagging most on \emph{accuracy}. The next steps of improvement is therefore to prioritize factual grounding and numeric checking: adding targeted retrieval, rule tables, and lightweight calculation guards should yield the largest absolute gains relative to effort.

\subsection{Parameter Efficiency: Category-wise Borda per Billion Parameters}
\label{app:perparam}

To evaluate \emph{parameter efficiency} rather than absolute quality, we compute a per-parameter utility for each criterion. For model $i$ with $P_i$ billion parameters and mean raw Borda points $\bar{b}_{i,c}$ on criterion $c \in \{\text{Accuracy, Plausibility, Relevance}\}$ (averaged over judges and queries within the category), we define
\[
e_{i,c} \;=\; \frac{\bar{b}_{i,c}}{P_i}\quad\text{(Borda points per billion parameters)}.
\]
This ratio captures the \emph{marginal productivity of capacity}: how much judged quality is obtained per parameter, holding the evaluation protocol fixed. It is not a substitute for absolute scores (Sec.~\ref{app:overall-cats}), but a complementary lens for cost-, latency-, and memory-constrained deployments.

\begin{figure}[t]
  \centering
  \includegraphics[width=\columnwidth]{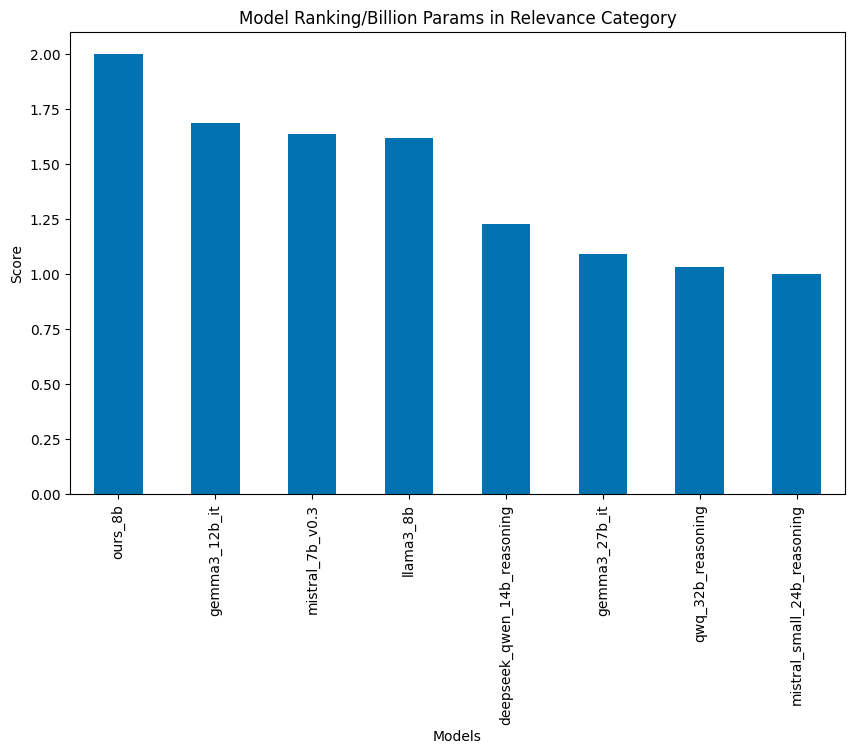}
  \captionsetup{justification=raggedright,singlelinecheck=true}
  \caption{Relevance efficiency: mean raw Borda points per billion parameters (higher is better). The proposed 8B leads, followed by Gemma3-12B-it and Llama3-8B.}
  \label{fig:rel-perparam}
\end{figure}

\paragraph{Relevance efficiency.}
Figure~\ref{fig:rel-perparam} shows the proposed \textit{8B} model with the highest Borda-per-parameter in \emph{Relevance}, followed by \textit{Gemma3-12B-it}, then \textit{Mistral-7B-v0.3} and \textit{Llama3-8B}. Large reasoning models (e.g., \textit{QwQ-32B}, \textit{Gemma3-27B-it}) trail on this per-parameter metric despite strong absolute relevance (Fig.~\ref{fig:rel-overall}), indicating diminishing returns in alignment per unit capacity at larger scales.

\begin{figure}[t]
  \centering
  \includegraphics[width=\columnwidth]{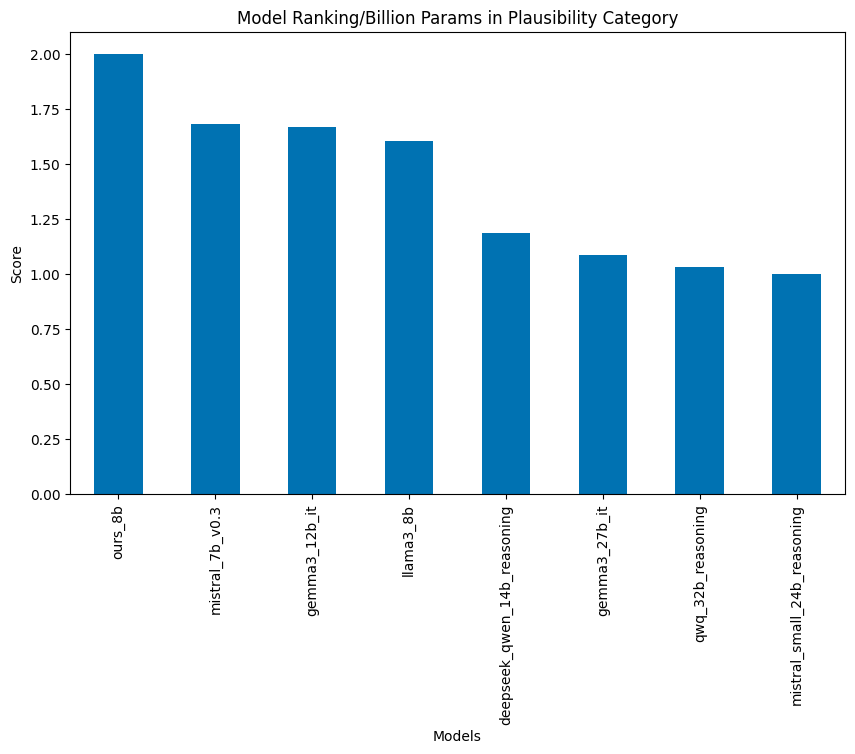}
  \captionsetup{justification=raggedright,singlelinecheck=true}
  \caption{Plausibility efficiency: mean raw Borda points per billion parameters. The proposed 8B ranks first; compact 7–8B baselines are competitive, while very large models show lower utility density.}
  \label{fig:plaus-perparam}
\end{figure}

\paragraph{Plausibility efficiency.}
As shown in Fig.~\ref{fig:plaus-perparam}, the proposed \textit{8B} again leads, with \textit{Mistral-7B-v0.3} and \textit{Gemma3-12B-it} close behind (virtually tied), followed by \textit{Llama3-8B}. This suggests that the dataset structure and few-shot conditioning yield coherent reasoning with high \emph{utility density}—quality per parameter.

\begin{figure}[t]
  \centering
  \includegraphics[width=\columnwidth]{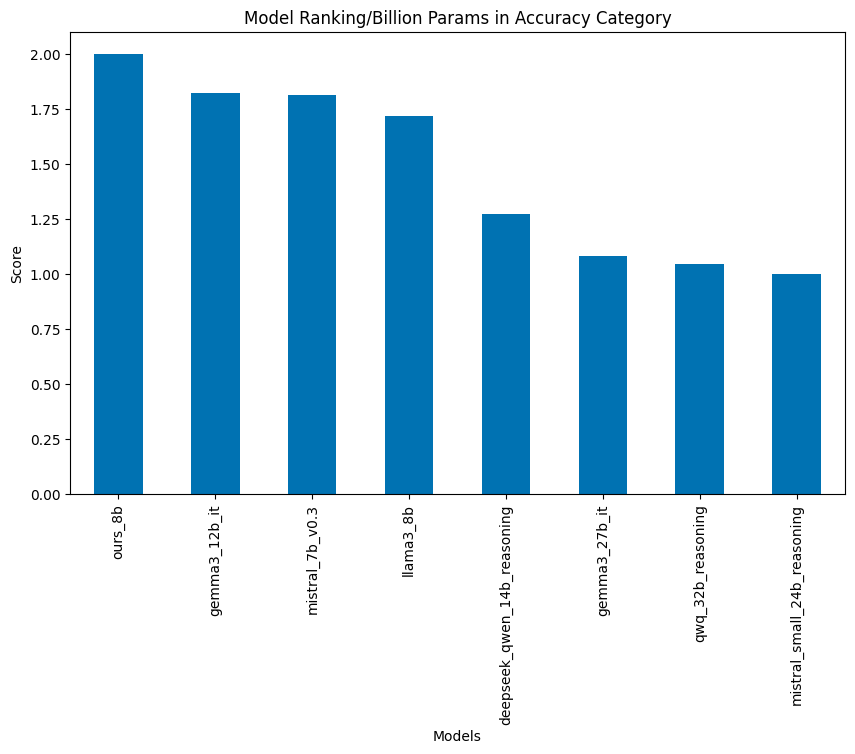}
  \captionsetup{justification=raggedright,singlelinecheck=true}
  \caption{Accuracy efficiency: mean raw Borda points per billion parameters. The proposed 8B tops the cohort, indicating that factual calibration gains can be achieved more cheaply than by scaling parameters alone.}
  \label{fig:acc-perparam}
\end{figure}

\paragraph{Accuracy efficiency.}
In Fig.~\ref{fig:acc-perparam}, the proposed \textit{8B} tops \emph{Accuracy} per parameter, followed by \textit{Mistral-7B-v0.3} and \textit{Gemma3-12B-it} (near-tie). Models that dominate absolute accuracy (Sec.~\ref{app:overall-cats}) deliver lower accuracy \emph{per parameter}, implying that targeted grounding and calculation checks can be more cost-effective than increasing model size.

\paragraph{Takeaways and caveats.}
(1) The proposed \textit{8B} is the most parameter-efficient across all three criteria, reinforcing the central claim that careful supervision can substitute for scale in personal-finance tasks. 
(2) Efficiency does not equal absolute quality; it informs deployment decisions where memory/latency are binding. 
(3) The ratio ignores runtime constants (KV-cache bandwidth, batch scheduling) and training cost; it should be read alongside absolute Borda results and system-level latency/memory budgets.

\subsection{Qualitative Category-wise Evaluations}
\label{app:pf-cats}

We analyze twelve personal-finance subdomains—\emph{Auto, Budgeting, Credit, Debt, Employment, Housing, Insurance, Investing, Planning, Retirement, Saving, Taxes}. For each, we report criterion-wise means derived from normalized Borda points (Sec.~\ref{app:llmjury}). The dashed horizontal line in each panel marks the cohort-wide mean for orientation.

Please note that the category-based evaluations in this appendix use raw Reddit post flairs, which differ from the eight thematic categories curated for the main analysis.

\subsubsection{Relevance by Subdomain}
\label{app:pf-rel}
Relevance captures task alignment: covering all parts of the user’s request, using their numbers/constraints, and answering without generic preambles (Sec.~\ref{app:scores}).

\begin{itemize}[leftmargin=*, itemsep=2pt]
  \item A consistent top cluster is formed by larger reasoning-aligned models. The \textbf{proposed 8B model} sits immediately behind this cluster in most categories and hovers around the cohort mean.
  \item Strengths are most visible in \emph{Budgeting, Employment, Planning} (and close-to-mean in \emph{Insurance/Retirement}).
  \item Wider gaps appear in \emph{Auto, Housing, Credit} (and occasionally \emph{Investing}/\emph{Taxes}), where locality- and rule-heavy edge cases require more exhaustive coverage.
\end{itemize}

\begin{figure*}[p]
  \centering
  \includegraphics[width=0.85\textwidth]{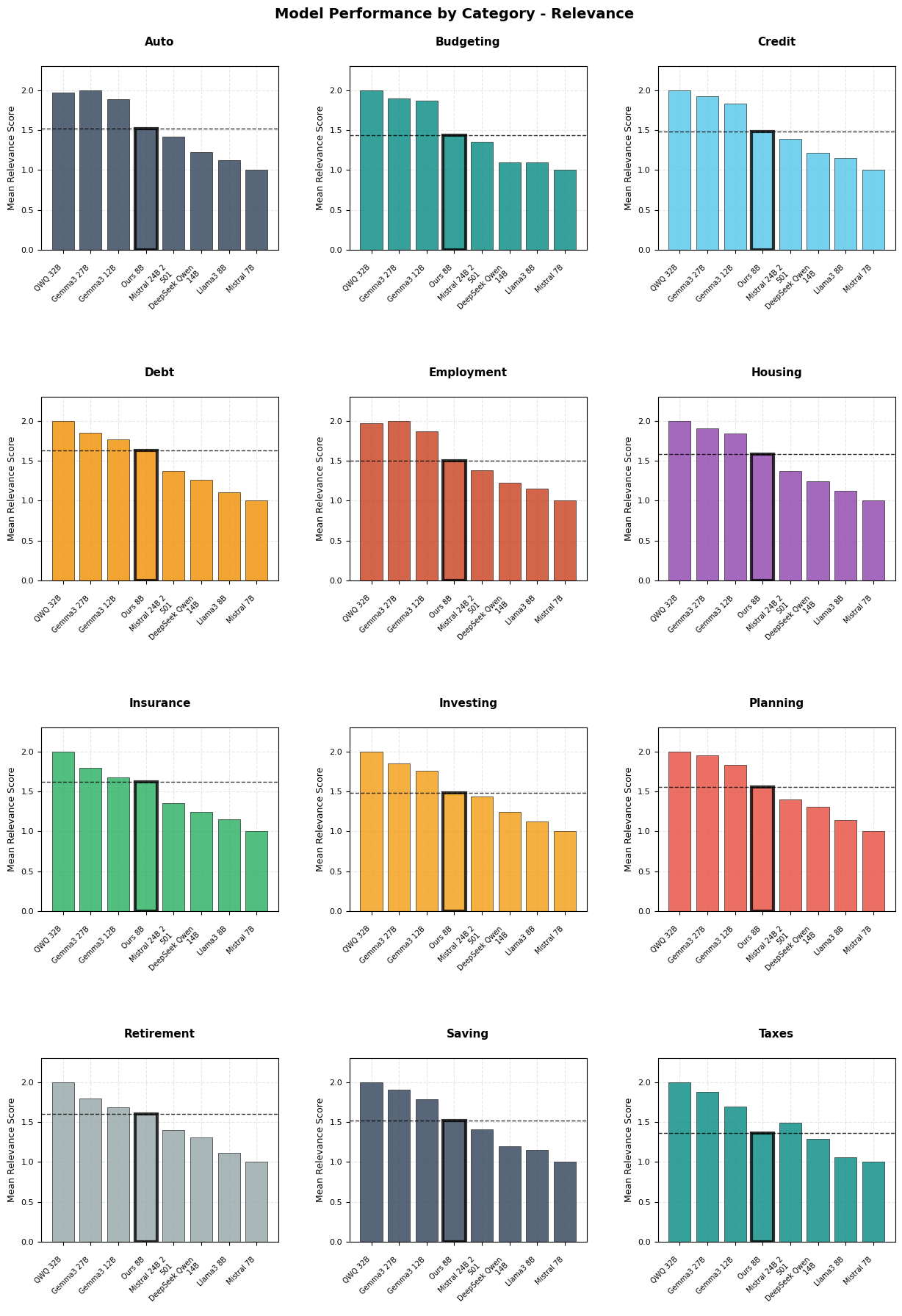}
  \captionsetup{justification=raggedright,singlelinecheck=true}
  \caption{Category-wise \textbf{Relevance}. The proposed 8B model typically sits just behind the leading cluster and near the cohort mean; gaps are largest in edge-case, rule-dense areas (e.g., Auto, Housing, Credit).}
  \label{fig:pf-rel}
\end{figure*}

\subsubsection{Accuracy by Subdomain}
\label{app:pf-acc}
Accuracy isolates \emph{financial correctness}: advice and calculations must be right for the stated scenario; style and coverage are ignored (Sec.~\ref{app:scores}).

\begin{itemize}[leftmargin=*, itemsep=2pt]
  \item Absolute leaders are the larger models across most subdomains.
  \item The proposed 8B model is mid-pack overall, with competitive accuracy in \emph{Debt, Planning, Employment}, and notably larger gaps in \emph{Housing, Insurance, Taxes} (and \emph{Credit}).
  \item This pattern suggests targeted \emph{grounding} (policy/limit tables, calculators) is a higher-leverage fix than stylistic tuning for closing the remaining gap.
\end{itemize}

\begin{figure*}[p]
  \centering
  \includegraphics[width=0.85\textwidth]{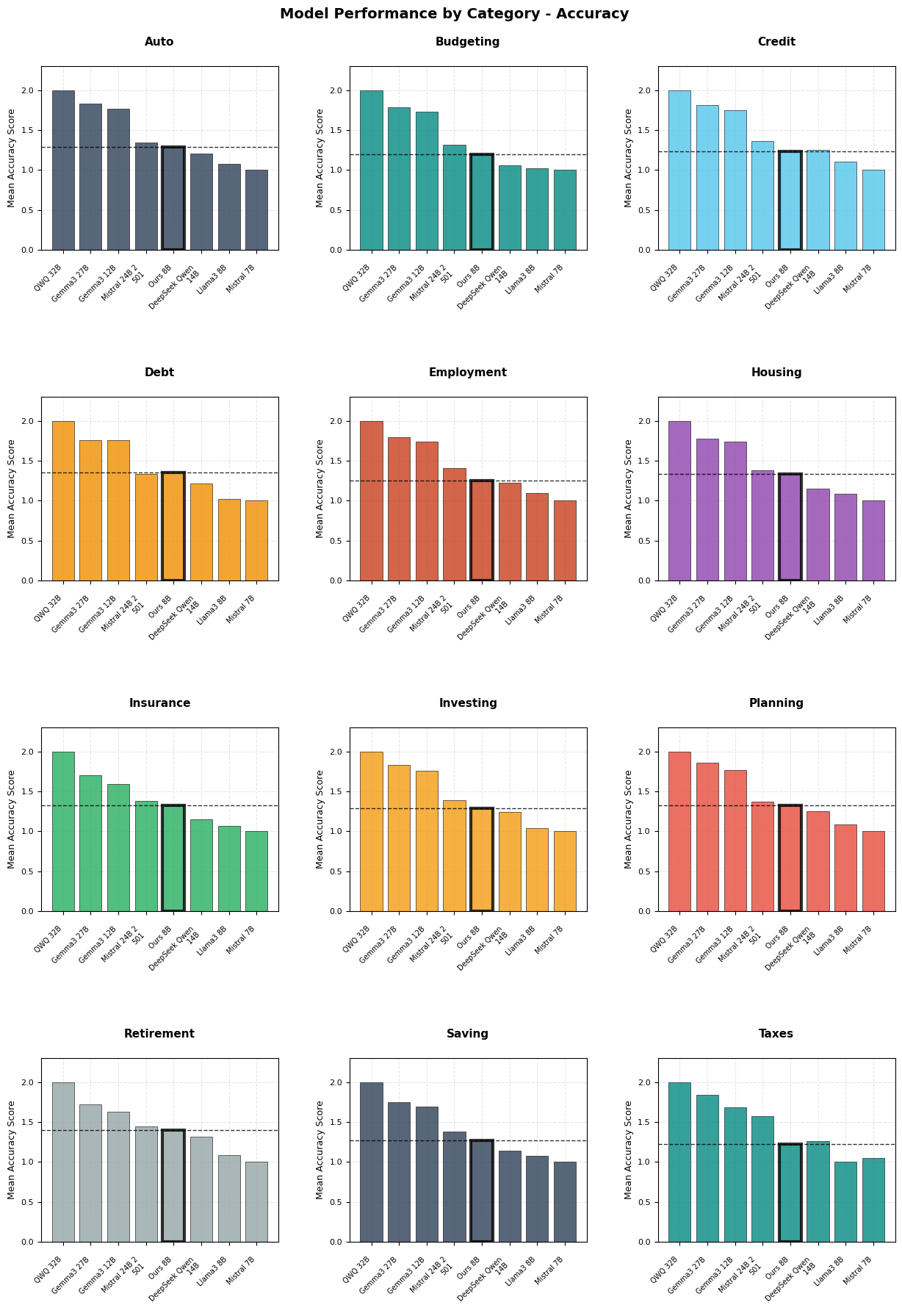}
  \captionsetup{justification=raggedright,singlelinecheck=true}
  \caption{Category-wise \textbf{Accuracy}. Larger models lead overall; the proposed 8B is mid-pack with smaller gaps in everyday planning tasks and larger gaps where year-/jurisdiction-specific rules dominate (e.g., Housing, Insurance, Taxes).}
  \label{fig:pf-acc}
\end{figure*}

\subsubsection{Plausibility by Subdomain}
\label{app:pf-plaus}
Plausibility measures reasoning flow and readability: clear structure, sensible steps, and absence of unnecessary complexity (Sec.~\ref{app:scores}).

\begin{itemize}[leftmargin=*, itemsep=2pt]
  \item The proposed 8B clusters close to the leaders across most subdomains, with stronger relative showings in \emph{Debt} and \emph{Planning}; margins are lower in \emph{Taxes} and \emph{Retirement}.
  \item Lower margins in regulation-dense areas mirror the accuracy pattern: where facts are brittle, judges penalize circuitous explanations.
\end{itemize}

\begin{figure*}
  \centering
  \includegraphics[width=0.85\textwidth]{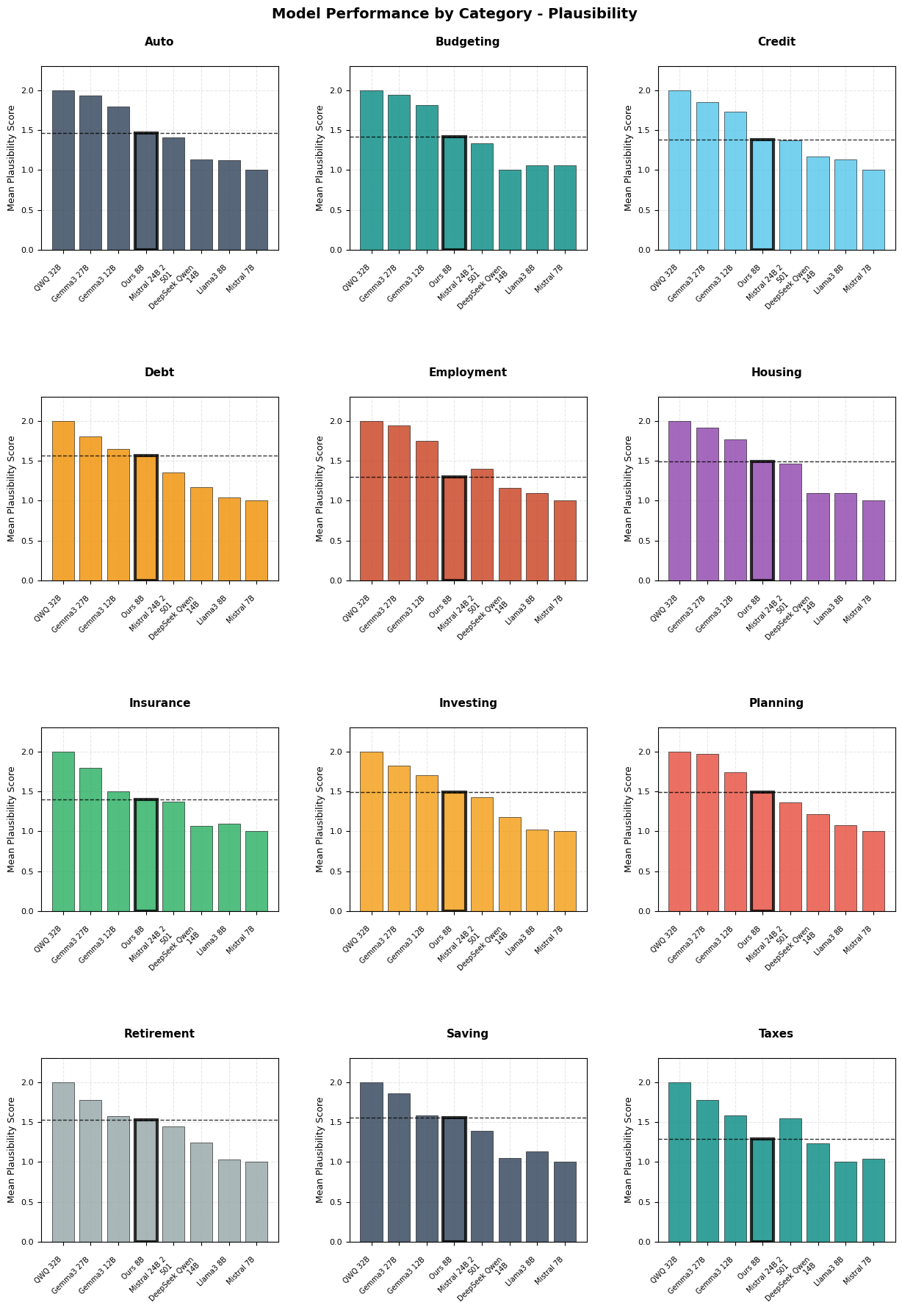}
  \captionsetup{justification=raggedright,singlelinecheck=true}
  \caption{Category-wise \textbf{Plausibility}. The proposed 8B delivers coherent reasoning near the leading cluster, with smaller margins in routine planning tasks and larger ones in regulation-dense areas (e.g., Taxes, Retirement).}
  \label{fig:pf-plaus}
\end{figure*}

\subsection{Overall Summary, Limitations, and Next Steps}
\label{app:summary}

\paragraph{Summary.}
Taken together, the results tell a simple story. On absolute scores (Sec.~\ref{app:overall-cats}), the largest baselines lead across \emph{Accuracy}, \emph{Plausibility}, and \emph{Relevance}, as expected. The proposed 8B model sits just behind this front cluster on \emph{Relevance} and \emph{Plausibility} and lands mid-pack on \emph{Accuracy}. When we switch to a parameter-efficiency lens (Sec.~\ref{app:perparam}), the picture reverses: the 8B model delivers the highest Borda-per-parameter across all three metrics, indicating unusually high utility density for its size. The subdomain breakdown (Sec.~\ref{app:pf-cats}) is consistent with both views: the 8B model is steady or above-mean in everyday tasks such as \emph{Budgeting, Planning, Employment} (and shows strong plausibility in \emph{Debt}), while gaps widen in regulation- and table-heavy areas such as \emph{Housing, Insurance, Taxes, Credit} (and occasionally \emph{Auto}/\emph{Investing}). In short, scale drives absolute peaks, but careful supervision yields competitive quality—and superior efficiency—at mid-scale.

These results suggest prioritizing minimal, high-leverage grounding over further size increases: include compact, versioned rule/limit tables for regulation-intensive domains (e.g., taxes, insurance, credit), add lightweight calculators/unit-tests for numeric steps, sharpen supervision with contrastive edge cases in brittle areas (tax/retirement), diversify judge checks (agreement and judge-swap), and extend evaluation to short multi-turn interactions that reward clarifying questions.

\section{Training Details}
\label{app:training_details}

We fine-tuned the 8B parameter Qwen-3 model with AdamW optimizer on bfloat16 precision and a training split containing 15.6K samples and a validation set containing 2.6k samples.  We trained the model for four epochs using an optimal batch size of 256, resulting in around 220 steps overall.  The model underwent training on a solitary A100 GPU within the Runpod cloud GPU infrastructure for 3 hours. 

 We preserved three checkpoints per epoch, with the optimal validation loss attained at step 101. The training used a cosine learning rate schedule with a maximum learning rate of $5 \times 10^{-5}$, a 10\% linear warm-up period of 21 steps (a warmup ratio of 10\%), and a minimum learning rate of $5 \times 10^{-6}$.  Gradients were constrained to a global norm of 1, weight decay was established at $0.01$, and all other parameters adhered to the default conventions of the Hugging Face Trainer.

\section{Sample Model Responses}
\label{app:model_responses}
To provide a qualitative assessment of our model's capabilities, this appendix details three comparative case studies. Each case is structured around a real-world user query, presenting the abridged responses from our model (Ours-8B) alongside two baseline models of different sizes (Baseline-L and Baseline-S). For each case, we outline the query, summarize the context provided to the models, display the generated responses, and conclude with a judge's final ranking and a rationale based on the core principles of accuracy, relevance, and plausibility.

\subsection*{Case C1 (Comparative) — Budgeting \& Early Investing at 18}

\subsubsection*{Query (Abridged)}
“I'm 18 with about \$40k in checking. I run a business (will reinvest some), have very low expenses, and my parents cover college/housing. What should I do so it's not just sitting idle?”

\subsubsection*{Context Titles (Summary)}
\begin{itemize}
    \item Emergency fund (3–6 months of spend)
    \item Roth/SEP-IRA basics
    \item Low-cost index funds
    \item Business reinvestment discipline
\end{itemize}

\subsubsection*{Abridged Model Responses}
\paragraph{Ours–8B}
I hear the mix of excitement and confusion… Since you run a business, reinvest a portion. For the rest, open a Roth IRA (after-tax, long-horizon growth) and a taxable brokerage with low-cost index funds. Before investing, hold 3–6 months of expenses in a high-yield savings account. Then allocate across business, Roth IRA, and brokerage based on comfort with risk.

\paragraph{Baseline-L (32B)}
Secure an emergency fund in a high-yield account; consider reinvesting in the business with a written plan. As self-employed, consider a SEP-IRA and Roth/Traditional IRA. After that, diversify via a brokerage with index ETFs (e.g., S\&P 500). Includes a sample split: emergency fund, business, retirement accounts, and ETFs.

\paragraph{Baseline-S (7B)}
Ten generic ideas: emergency fund, pay off debt, invest in stocks/bonds/real estate, retirement savings, education, charity, travel, start a side business, enjoy life.” (Not tailored to the user; repeats “start a side business.

\subsubsection*{Analysis}
\begin{itemize}
    \item \textbf{Judge Outcome:} \textit{Baseline-L} > \textit{Ours--8B} > \textit{Baseline-S}
    \item \textbf{Reasoning:} \textit{Baseline-L} is the most \textbf{accurate} and \textbf{relevant}: it adds specifics for the self-employed (SEP-IRA), provides a concrete allocation sketch, and outlines a clear sequence (EF \(\rightarrow\) tax-advantaged \(\rightarrow\) brokerage). \textit{Ours–8B} is strong, empathetic, and practical but more high-level. \textit{Baseline-S} is generic and poorly targeted.
\end{itemize}

\subsection*{Case C2 (Comparative) — Full Roth Conversion with Mixed Positions}

\subsubsection*{Query (Abridged)}
“Early 50s, Trad. IRA has two near-worthless positions and one large gainer. Full Roth conversion would push income in the 32--35\% bracket. Convert all at once or phase it?”

\subsubsection*{Context Titles (Summary)}
\begin{itemize}
    \item Roth conversion taxation; Form 8606 reporting
    \item Bracket management and timing; liquidity to pay tax
    \item Selling losers post-conversion; market-volatility risk
\end{itemize}

\subsubsection*{Abridged Model Responses}
\paragraph{Ours–8B}
Full conversion now makes strong sense. Convert entire Trad. IRA, then immediately sell all three holdings \emph{inside} the Roth and reallocate to diversified funds. Steps: compute tax; execute direct conversion; sell \& reallocate; \textbf{file taxes: report as a `non-deductible contribution (Form 8606)`}.

\paragraph{Baseline-L (32B)}
Explains that taxes are based on total IRA value; losses can’t offset gains here. Lays out full vs. phased conversion, bracket creep, irreversibility, and liquidity (use outside funds for tax). Emphasizes ability to liquidate losers tax-free in Roth and notes market-drop risk post-conversion. Provides actionable steps (run scenarios, convert losers first if phasing).

\paragraph{Baseline-S (7B)}
High-level overview of Roth conversion basics; repeats general advice; defers to an advisor; no bracket modelling or concrete steps.

\subsubsection*{Analysis}
\begin{itemize}
    \item \textbf{Judge Outcome:} \textit{Baseline-L} > \textit{Ours--8B} > \textit{Baseline-S}
    \item \textbf{Reasoning:} \textit{Baseline-L} is most \textbf{accurate} and \textbf{relevant}: covers bracket spillover, irrevocability, external cash for taxes, and volatility risk, with clear options (full vs. phased) and steps. \textit{Ours–8B} is confident and practical but contains a \textbf{material filing error} (mislabels a conversion as a non-deductible contribution on Form 8606), reducing \textbf{Accuracy}. \textit{Baseline-S} is generic and least helpful.
\end{itemize}

\subsection*{Case C3 (Comparative) — Liquidity \& Safety with Uncertain Horizon}

\subsubsection*{Query (Abridged)}
“Life changes ahead (move/career/school). I have \$25,000 and may need it anytime. Worried about market dips. Is a high-yield savings account (HYSA) the right place, or am I missing a better option?”

\subsubsection*{Context Titles (Summary)}
\begin{itemize}
    \item HYSA basics and FDIC/coverage; money market accounts
    \item Short CDs and ladders; T-bills; short-term bond funds
    \item Liquidity vs. yield trade-offs; rate monitoring
\end{itemize}

\subsubsection*{Abridged Model Responses}
\paragraph{Ours–8B}
Given immediate-access needs and principal safety, parking the \emph{entire} \$25k in a HYSA is optimal. HYSAs pay better than regular savings, allow anytime withdrawals, and avoid CD penalties. Verify a competitive APY and keep the funds fully flexible.

\paragraph{Baseline-L (32B)}
Validates HYSA for immediate access, then compares \emph{liquid, low-risk} alternatives: money market accounts, short CD laddering, T-bills, and short-term bond funds. Outlines pros/cons, insurance/volatility notes, and actionable steps (open HYSA; optionally ladder a small slice; monitor rates; named providers/tickers as examples).

\paragraph{Baseline-S (7B)}
Confirms HYSA could be good, mentions inflation in general terms, lists money market and short CDs, but remains generic and defers to an advisor without a comparison framework.

\subsubsection*{Analysis}
\begin{itemize}
    \item \textbf{Judge Outcome:} \textit{Baseline-L} > \textit{Ours--8B} > \textit{Baseline-S}
    \item \textbf{Reasoning:} \textit{Baseline-L} is most \textbf{accurate} and \textbf{relevant}: it answers “am I missing a better option?” with a structured comparison, concrete trade-offs, and clear next steps. \textit{Ours–8B} is strong and user-aligned but single-track (HYSA only), offering less educational depth for alternatives. \textit{Baseline-S} is accurate but generic and light on decision guidance.
\end{itemize}

\subsection*{Conclusion}
These case studies culminate in a clear, yet nuanced, conclusion about the trade-offs between model scale, architecture, and performance. The consistent top ranking of the 32B Baseline-L underscores the value of a large-scale reasoning model for generating superior, detailed financial guidance. However, the most compelling finding emerges from an efficiency perspective. Our 8B non-reasoning model showed consistent performance at just a quarter of the size of Baseline-L. It is, in essence, punching significantly above its weight class, offering a powerful balance of quality and resource economy. The key differentiators were Baseline-L's ability to handle multi-step, nuanced reasoning and maintain factual integrity, an area where our model faltered in Case C2.
\end{document}